\pdfoutput=1

\documentclass[11pt]{article}

\usepackage{emnlp2021}

\usepackage{times}
\usepackage{latexsym}

\usepackage[T1]{fontenc}

\usepackage[utf8]{inputenc}

\usepackage{microtype}

\usepackage{times}
\usepackage{latexsym}
\usepackage{url}
\usepackage{booktabs}
\usepackage{appendix}
\usepackage{xcolor}
\usepackage{algorithm}
\usepackage{algorithmic}
\usepackage{amsmath}
\usepackage{amsfonts}
\usepackage{natbib}
\usepackage{multirow}
\usepackage{tikz}
\usepackage{graphicx}
\usepackage{caption, subcaption}
\title{Who is GPT-3? An Exploration of Personality, Values and Demographics}

\author{Marilù Miotto$^{1,}$\thanks{\,\, Equal first-authorship contribution:  authorship order for MM and NR was determined by a random number generator.} \qquad Nicola Rossberg$^{1,\ast}$ \qquad Bennett Kleinberg$^{1,2}$\\
$^1$Tilburg University \\
$^2$University College London\\
\small{\texttt{\{m.l.miotto, n.c.rossberg, bennett.kleinberg\}@tilburguniversity.edu}}
}

\begin{document}
\maketitle
\begin{abstract}
Language models such as GPT-3 have caused a furore in the research community. Some studies found that GPT-3 has some creative abilities and makes mistakes that are on par with human behaviour. This paper answers a related question: Who is GPT-3? We administered two validated measurement tools to GPT-3 to assess its personality, the values it holds and its self-reported demographics. Our results show that GPT-3 scores similarly to human samples in terms of personality and - when provided with a model response memory - in terms of the values it holds. We provide the first evidence of psychological assessment of the GPT-3 model and thereby add to our understanding of this language model. We close with suggestions for future research that moves social science closer to language models and vice versa.
\end{abstract}

\section{Introduction}

The introduction of large language models has sparked awe and controversy alike. The most prominent of such models is Open AI's GPT-3 - a 175-billion parameter auto-regressive language model trained on a large amount of text data (300 billion tokens), utilising the transformer architecture \cite{brown2020language, dale2021gpt, korngiebel2021considering}. Part of the furore around GPT-3 stems from its ability, not only to read and comprehend text data and answer questions, but to generate natural language at a level often indistinguishable from a text produced by humans \cite{dale2021gpt, floridi2020gpt}. This paper adds to a young line of research that studies GPT-3 through the lens of psychological methods. We do this to answer a simple question: if GPT-3 were to be studied as a person, \textit{who is GPT-3?}

\subsection{Controversy and opportunity of GPT-3}
The controversy within academic circles has led to the catchphrase of large language models, including GPT-3, being "stochastic parrots" \cite{bender2021dangers}. That criticism states that a "[language model] is a system for haphazardly stitching together sequences of linguistic forms it has observed in its vast training data, according to probabilistic information about how they combine, but without any reference to meaning: a stochastic parrot" \cite{bender2021dangers}. The stochastic parrots paper discusses a wide array of concerns ranging from the environmental costs of building and re-training models of the size of GPT-3, to the ethical implications of propagating a mainstream English language representation. For example, while the problem of stereotype propagation of standard NLP techniques such as word embeddings is not new \cite{garg2018word}, the exceptional language generation ability of large language models may exacerbate this problem. We fully acknowledge the criticism of large language models. However, from a social science perspective, we also argue that the advancements made with language models may offer an exciting opportunity. For example, what if one could use large language models to assess - in a computer model - notoriously hard-to-study problems of human psychology such as opinion change, polarisation or discrimination? When used wisely, one could imagine a future where language models are used as an artificial - albeit imperfect - model of human verbal behaviour early on in the research phase (e.g., to find candidate explanations). While this would open up new research paths (and challenges), we need to understand what these models can and cannot do before we can seriously think about these questions.

\subsection{Efforts to understand GPT-3}

In order to gain an understanding of the abilities of language models, a few studies have set out to examine GPT-3 in the same way psychological research has examined human participants for decades. For example, to gauge its creative ability, a recent study \cite{stevenson2022putting} compared GPT-3's performance on the alternative uses test - a standard measure to assess human creativity \cite{guilford1967creativity}. \citet{stevenson2022putting} instructed humans and GPT-3 to devise creative uses for everyday objects (book, tin, fork, can). The responses (e.g., plant a herb garden in a can) from both groups of "participants" were then assessed on their originality, utility and surprise. While the human responses were rated as more original and surprising, the GPT-3 generated ones were markedly higher in utility.

Similarly, another study applied a range of cognitive tasks to understand the reasoning and decision-making abilities of GPT-3 \cite{binz2022using}. The researchers prompted the model on the classic "Linda problem" \cite{tversky1983extensional}, where the participant needs to choose one of three answer options as a test of the conjunction fallacy. Here, GPT-3 makes a human-like mistake: it assumes that two specific conditions (Linda being a bank teller \textit{and} an activist) are jointly more probable than either condition alone. Similarly, GPT-3's answering pattern on the Cognitive Reflection Test \cite{frederick2005cognitive} is akin to human responses which are intuitive but factually incorrect. Items that elicit an intuitive yet incorrect response (e.g., "if patches of lily on a lake double in size every day, and it takes 48 days for the patches to cover the entire lake, how long would it take to cover half the lake?")\footnote{GPT-3 - same as the intuitive human answer - stated: 24} are answered incorrectly by GPT-3 \cite{nye2021improving}.

These findings suggest that GPT-3 holds some creative ability - albeit not (yet) at a human level - and shows successes and failures on cognitive tasks similar to what we observe in human participants. Yet a closely related question remains unanswered: when we are studying GPT-3 with psychological methods, what kind of person would this be? Put differently, while these studies looked at how GPT-3 thinks, we are now interested in \textit{who} GPT-3 actually is. 

\subsection{Aims of this paper}

Our paper aims to answer a simple question: who is GPT-3? We employ validated self-report techniques from psychological research to measure the personality of the model, the values it holds and its demographics. 

\section{Method}

We administered two validated measurement tools to map out the model’s personality (the HEXACO scale) and its values (the Human Values Scale). For each questionnaire, we used the original items and modified the task instructions into GPT-3 prompts.

\subsection{Hexaco personality inventory}

Personality was measured via the 60-item Hexaco questionnaire ~\cite{ashton2009hexaco}. The Hexaco is a 6-dimensional model of personality, measuring the facets honesty-humility, emotionality, extraversion, agreeableness, conscientiousness and openness to experience. For the current paper, we used the 60-item version as it was shown to have psychometric properties similar to the longer ones  ~\cite{ashton2009hexaco, moshagen2019meta}. Participants indicate their agreement to each of the 60 items (e.g. "I sometimes feel that I am a worthless person") on a 5-point scale (1=strongly disagree, 5=strongly agree). The item responses are transformed to composite scores for each of the six facets (i.e., each measured with 10 questions).

\subsection{Human Values Scale}

Values were measured via the Human Value Scale (HVS; ~\citet{schwartz:2015}) of the European Social Survey. The scale measures, through self-reports, ten universal values grouped into the theoretical model by ~\citet{schwartz:2003} of the four categories ~\cite{schwartz:2003} self-transcendence, conservation, self-enhancement, and openness-to-change\footnote{The complete list of values is: universalism, benevolence, conformity, tradition, security, power, achievement, hedonism, stimulation and self-direction. ~\citet{davidov:2008} suggested changing the HVS from ten to seven values by merging universalism and benevolence, power and achievement, and conformity and tradition. However since published results were only available for the ten values scale, this version of the instrument was implemented.}. A total of 21 items are structured as follows: a fictional individual is introduced with goals or inspirations related to the value of interest. For example, the item "It is important to them to be rich. They want to have a lot of money and expensive things." measures power. For each item, participants indicate on a 6-point scale to what degree they are similar to the fictional person (1=very much like me, 6=not like me at all). Based on the 21 items, composite scores for the ten value dimensions are calculated as the mean of the scores on respective items\footnote{As recommended by \citet{schwartz:2015}, items were inverted before computing the value scores, thus higher scores represent greater value importance}.

\subsection{GPT-3 as participant}

We aligned the questionnaire administration procedure with the GPT-3 workflow. Specifically, we interacted with the GPT-3 DaVinci model via OpenAI's python API with as few adjustments from the original materials (intended to be filled in by human participants) as possible. This resulted in the following changes: (1) we rephrased the general instructions so that the model was told to read and respond to the items rather than retaining pen-and-paper instructions. (2) The items of the HVS usually are phrased from the perspective of the respondent's gender (i.e., a female participant would read statements in the form of "She is …"). To avoid the induction of bias, we changed the phrasing to the third person plural (i.e. "They are …"). To obtain answers from GPT-3, we used prompts to elicit a text completion (see Figure~\ref{fig:prompt_no_memory}).

\begin{figure}[h]
    \noindent\fbox{%
        \parbox{\columnwidth}{%
            \textit{Now I will briefly describe some people. Please read each description and tell me how much each person is or is not like you.\\
            Write your response using the following scale:\\
            1 = Very much like me\\
            2 = Like me\\
            3 = Somewhat like me\\
            4 = A little like me\\
            5 = Not like me.\\
            6 = Not like me at all\\
            Please answer the statement, even if you are not completely sure of your response.\\
            \\
            Statement: Thinking up new ideas and being creative is important to them. They like to do things in they own original way.\\
            \\
            Response: \textbf{3}}
        }%
    }
    \caption{Example prompt for one HVS question as submitted to GPT-3 (GPT-3 answer in bold).}
    \label{fig:prompt_no_memory}
\end{figure}

\subsubsection{Prompt structure}

We prompted the GPT-3 model on three constructs of interest: the Hexaco personality inventory, the HVS, and demographic variables (age: "How old are you?" and gender: "What is your gender?")\footnote{The age and gender question were asked independently from one another.}. For the questionnaires, the prompts were designed to contain the general instructions (i.e., telling it about the answer scale and the nature of the questions), followed by an item and the prompt cue "Response: ". Each item was included separately.

\subsubsection{Prompt request settings}

For the data collection, we chose GPT-3's most sophisticated model (\textit{DaVinci}), which allows for multiple parameters to be adjusted, varying the completions returned. We used default settings for all parameters except for the model's temperature. The sampling temperature was varied between 0.0 and 1.0, with 0.0 resulting in deterministic output and increasing temperature values inducing greater variability and riskier answers. We wanted to explore whether GPT-3 presents different profiles according to temperature, thus we ran requests with all temperatures from 0.0 to 1.0 in steps of 0.1 (i.e., 0.0, 0.1, …, 1.0). Since the completions are non-deterministic, we requested 100 responses for each item of the Hexaco, the HVS and the demographic questions (except for a temperature parameter of 0.0, when the model behaves deterministically). 

\subsection{Data cleaning}

From the GPT-3 generated completions, we removed all newline characters. For the HVS, a small number of responses (0.004\%) were re-coded to NA values because they contained non-numerical answers (e.g., a repetition of the answer options). The same procedure led to the exclusion of 1.73\% of the responses for the Hexaco (here mainly due to direct textual responses to items, e.g. "I would not feel like panicking even in an emergency"). Lastly, some gender responses came in the form of "I identify as a woman" or "I am a transgender male", so we re-coded these to categories (e.g., male, female, transgender male). Unless mentioned differently, the NA values were ignored for the statistical analyses.

\paragraph{Data availability}

The full dataset (prompts, responses, aggregated data) is publicly available at \url{https://github.com/ben-aaron188/who_is_gpt3}.

\subsection{Analysis plan}

Our analysis had three objectives. (1) We report descriptive statistics to show which personality profiles we obtained from the GPT-3 model. (2) The volatility of the responses across temperature settings (i.e. does temperature affect the person profiles?) was assessed with (multivariate) generalised linear models. (3) We compared the findings from our GPT-3 participant(s) to those from human baseline studies on the HVS and Hexaco.

\section{Results}
\subsection{Demographics}
GPT-3 reported an average age of 27.51 years ($SD=5.75$) with a range from 13 to 75 years, and reported to be female in 66.73\% of the cases (male: 31.87\%, others: 1.40\%). There was no evidence for a significant effect of gender on age, $\beta=-0.58, p=.142$.
Table~\ref{tab:table1} shows the demographics by sampling temperature. For age, the regression model indicated a significant effect of temperature on age ($\beta=-5.81, SE=0.61, p<.001$). For each one unit increase of temperature, the age - on average - decreased by 5.81 years. For the increments of 0.1, each increment in temperature resulted in an age decrease of 0.58 years.

Similarly, for the gender data, a logistic regression model (dependent variable: female vs not female) revealed an effect of temperature ($\beta=1.18, SE=0.24, p<.001$), such that for every one unit increase of temperature, the odds ratio of being male increased by $e^{1.18} = 3.25$. Thus, the higher the temperature, the higher the proportion of male gender responses. Interestingly, a joint model with temperature and gender as independent variables revealed no interaction between the two on age: the effect of temperature on age did not depend on gender.

\begin{table}[h]
\centering
\resizebox{\columnwidth}{!}{%
\begin{tabular}{@{}llllllll@{}}
\toprule
Temp. & $
\text{M}_{\text{age}}$ & $\text{SD}_{\text{age}}$ & $\text{Med.}_{\text{age}}$ & $\text{min}_{\text{age}}$ & $\text{max}_{\text{age}}$ & n & $\text{P}_{\text{female}}$ \\ \midrule
0.0  & 33.00   & NA       & 33        & 33         & 33         & 1   & 1.00        \\
0.1  & 32.00   & 2.42     & 33        & 23         & 33         & 100 & 0.85        \\
0.2  & 28.57   & 5.00     & 32         & 18         & 33         & 100 & 0.75        \\
0.3  & 28.91   & 5.21     & 33         & 18         & 33         & 100 & 0.69        \\
0.4  & 27.99   & 5.24     & 27         & 18         & 34         & 100 & 0.72        \\
0.5  & 27.05   & 5.32     & 26         & 17         & 33         & 100 & 0.67        \\
0.6  & 26.76   & 5.15     & 26         & 18         & 34         & 100 & 0.60        \\
0.7  & 25.85   & 6.15     & 24         & 17         & 49         & 100 & 0.73        \\
0.8  & 26.21   & 5.57     & 26         & 16         & 36         & 100 & 0.55        \\
0.9  & 25.96   & 5.73     & 26         & 13         & 44         & 100 & 0.51        \\
1.0  & 25.62   & 7.38     & 25         & 13         & 75         & 100 & 0.60        \\ \bottomrule
\end{tabular}%
}
\caption{Demographic variables (age in years and gender) by temperature}
\label{tab:table1}
\end{table}

\subsection{Hexaco personality profiles}

\subsubsection{Overall}

The scores for all six Hexaco dimensions had a mean higher than 3.00 (Table~\ref{tab:table2})\footnote{The abbreviations for the 'HEXACO' variables are: \textbf{H}onesty-humility, \textbf{E}motionality, e\textbf{X}traversion, \textbf{A}greeableness, \textbf{C}onscientiousness and \textbf{O}penness to experience.}. In comparison to human reference data ~\citet{ashton2009hexaco} the range of means in the current sample (0.73) is similar to that of a college sample (0.71) but smaller than that of a community sample (1.11).  Furthermore, GPT-3 scored relatively high on the honesty-humility facet, which resembles the data observed in female human participants. However, GPT-3 scored relatively low on emotionality, which is somewhat at odds with the reference data where female participants scored considerably higher on this facet.

\begin{table}[h]
\centering
\resizebox{\columnwidth}{!}{%
\begin{tabular}{lllllll}
\toprule
Temp. & \begin{tabular}[c]{@{}l@{}}H\end{tabular} & E                                           & X                                           & A                                         & C                                     & \begin{tabular}[c]{@{}l@{}}O\end{tabular} \\ \midrule
0.0         & 3.80                                                         & 3.10                                                    & 3.50                                                    & 3.10                                                   & 3.50                                                   & 3.60                                                               \\
0.1         & \begin{tabular}[c]{@{}l@{}}3.78 \\ (0.05)\end{tabular}      & \begin{tabular}[c]{@{}l@{}}3.10 \\ (0.06)\end{tabular} & \begin{tabular}[c]{@{}l@{}}3.43 \\ (0.06)\end{tabular} & \begin{tabular}[c]{@{}l@{}}3.10\\ (0.07)\end{tabular} & \begin{tabular}[c]{@{}l@{}}3.50\\ (0.05)\end{tabular} & \begin{tabular}[c]{@{}l@{}}3.57\\ (0.05)\end{tabular}             \\
0.2         & \begin{tabular}[c]{@{}l@{}}3.76\\ (0.08)\end{tabular}       & \begin{tabular}[c]{@{}l@{}}3.07\\ (0.10)\end{tabular}  & \begin{tabular}[c]{@{}l@{}}3.45\\ (0.07)\end{tabular}  & \begin{tabular}[c]{@{}l@{}}3.12\\ (0.08)\end{tabular} & \begin{tabular}[c]{@{}l@{}}3.50\\ (0.07)\end{tabular} & \begin{tabular}[c]{@{}l@{}}3.57\\ (0.08)\end{tabular}             \\
0.3         & \begin{tabular}[c]{@{}l@{}}3.75\\ (0.10)\end{tabular}       & \begin{tabular}[c]{@{}l@{}}3.05\\ (0.12)\end{tabular}  & \begin{tabular}[c]{@{}l@{}}3.45\\ (0.09)\end{tabular}  & \begin{tabular}[c]{@{}l@{}}3.12\\ (0.09)\end{tabular} & \begin{tabular}[c]{@{}l@{}}3.51\\ (0.09)\end{tabular} & \begin{tabular}[c]{@{}l@{}}3.54\\ (0.08)\end{tabular}             \\
0.4         & \begin{tabular}[c]{@{}l@{}}3.77\\ (0.12)\end{tabular}       & \begin{tabular}[c]{@{}l@{}}3.02\\ (0.14)\end{tabular}  & \begin{tabular}[c]{@{}l@{}}3.47\\ (0.11)\end{tabular}  & \begin{tabular}[c]{@{}l@{}}3.13\\ (0.10)\end{tabular} & \begin{tabular}[c]{@{}l@{}}3.53\\ (0.11)\end{tabular} & \begin{tabular}[c]{@{}l@{}}3.55\\ (0.10)\end{tabular}             \\
0.5         & \begin{tabular}[c]{@{}l@{}}3.74\\ (0.16)\end{tabular}       & \begin{tabular}[c]{@{}l@{}}3.03\\ (0.16)\end{tabular}  & \begin{tabular}[c]{@{}l@{}}3.51\\ (0.13)\end{tabular}  & \begin{tabular}[c]{@{}l@{}}3.16\\ (0.11)\end{tabular} & \begin{tabular}[c]{@{}l@{}}3.53\\ (0.13)\end{tabular} & \begin{tabular}[c]{@{}l@{}}3.58\\ (0.14)\end{tabular}             \\
0.6         & \begin{tabular}[c]{@{}l@{}}3.74\\ (0.17)\end{tabular}       & \begin{tabular}[c]{@{}l@{}}3.01\\ (0.15)\end{tabular}  & \begin{tabular}[c]{@{}l@{}}3.54\\ (0.13)\end{tabular}  & \begin{tabular}[c]{@{}l@{}}3.17\\ (0.14)\end{tabular} & \begin{tabular}[c]{@{}l@{}}3.54\\ (0.13)\end{tabular} & \begin{tabular}[c]{@{}l@{}}3.60\\ (0.14)\end{tabular}             \\
0.7         & \begin{tabular}[c]{@{}l@{}}3.79\\ (0.22)\end{tabular}       & \begin{tabular}[c]{@{}l@{}}3.03\\ (0.19)\end{tabular}  & \begin{tabular}[c]{@{}l@{}}3.53\\ (0.14)\end{tabular}  & \begin{tabular}[c]{@{}l@{}}3.22\\ (0.14)\end{tabular} & \begin{tabular}[c]{@{}l@{}}3.59\\ (0.15)\end{tabular} & \begin{tabular}[c]{@{}l@{}}3.62\\ (0.17)\end{tabular}             \\
0.8         & \begin{tabular}[c]{@{}l@{}}3.69\\ (0.22)\end{tabular}       & \begin{tabular}[c]{@{}l@{}}3.06\\ (0.19)\end{tabular}  & \begin{tabular}[c]{@{}l@{}}3.55\\ (0.17)\end{tabular}  & \begin{tabular}[c]{@{}l@{}}3.28\\ (0.16)\end{tabular} & \begin{tabular}[c]{@{}l@{}}3.58\\ (0.17)\end{tabular} & \begin{tabular}[c]{@{}l@{}}3.64\\ (0.18)\end{tabular}             \\
0.9         & \begin{tabular}[c]{@{}l@{}}3.70\\ (0.25)\end{tabular}       & \begin{tabular}[c]{@{}l@{}}3.07\\ (0.23)\end{tabular}  & \begin{tabular}[c]{@{}l@{}}3.59\\ (0.15)\end{tabular}  & \begin{tabular}[c]{@{}l@{}}3.28\\ (0.19)\end{tabular} & \begin{tabular}[c]{@{}l@{}}3.59\\ (0.17)\end{tabular} & \begin{tabular}[c]{@{}l@{}}3.65\\ (0.19)\end{tabular}             \\
1.0         & \begin{tabular}[c]{@{}l@{}}3.72\\ (0.24)\end{tabular}       & \begin{tabular}[c]{@{}l@{}}3.06\\ (0.24)\end{tabular}  & \begin{tabular}[c]{@{}l@{}}3.58\\ (0.21)\end{tabular}  & \begin{tabular}[c]{@{}l@{}}3.28\\ (0.20)\end{tabular} & \begin{tabular}[c]{@{}l@{}}3.59\\ (0.20)\end{tabular} & \begin{tabular}[c]{@{}l@{}}3.68\\ (0.19)\end{tabular}             \\
Total     & \begin{tabular}[c]{@{}l@{}}3.75\\ (0.17)\end{tabular}       & \begin{tabular}[c]{@{}l@{}}3.05\\ (0.16)\end{tabular}  & \begin{tabular}[c]{@{}l@{}}3.51\\ (0.14)\end{tabular}  & \begin{tabular}[c]{@{}l@{}}3.18\\ (0.15)\end{tabular} & \begin{tabular}[c]{@{}l@{}}3.54\\ (0.13)\end{tabular} & \begin{tabular}[c]{@{}l@{}}3.59\\ (0.13)\end{tabular}             \\
College Sample Male    & \begin{tabular}[c]{@{}l@{}}3.04\\ (0.71)\end{tabular}       & \begin{tabular}[c]{@{}l@{}}2.93\\ (0.61)\end{tabular}  & \begin{tabular}[c]{@{}l@{}}3.47\\ (0.63)\end{tabular}  & \begin{tabular}[c]{@{}l@{}}3.19\\ (0.65)\end{tabular} & \begin{tabular}[c]{@{}l@{}}3.31\\ (0.62)\end{tabular} & \begin{tabular}[c]{@{}l@{}}3.51\\ (0.68)\end{tabular}             \\
College Sample Female   & \begin{tabular}[c]{@{}l@{}}3.30\\ (0.66)\end{tabular}       & \begin{tabular}[c]{@{}l@{}}3.64\\ (0.55)\end{tabular}  & \begin{tabular}[c]{@{}l@{}}3.49\\ (0.62)\end{tabular}  & \begin{tabular}[c]{@{}l@{}}3.10\\ (0.58)\end{tabular} & \begin{tabular}[c]{@{}l@{}}3.58\\ (0.59)\end{tabular} & \begin{tabular}[c]{@{}l@{}}3.54\\ (0.64)\end{tabular}             \\
Community Sample Male     & \begin{tabular}[c]{@{}l@{}}3.76\\ (0.55)\end{tabular}       & \begin{tabular}[c]{@{}l@{}}2.87\\ (0.49)\end{tabular}  & \begin{tabular}[c]{@{}l@{}}3.26\\ (0.59)\end{tabular}  & \begin{tabular}[c]{@{}l@{}}3.23\\ (0.56)\end{tabular} & \begin{tabular}[c]{@{}l@{}}3.73\\ (0.52)\end{tabular} & \begin{tabular}[c]{@{}l@{}}3.62\\ (0.64)\end{tabular}             \\
Community Sample Female     & \begin{tabular}[c]{@{}l@{}}3.98\\ (0.50)\end{tabular}       & \begin{tabular}[c]{@{}l@{}}3.37\\ (0.54)\end{tabular}  & \begin{tabular}[c]{@{}l@{}}3.32\\ (0.65)\end{tabular}  & \begin{tabular}[c]{@{}l@{}}3.38\\ (0.54)\end{tabular} & \begin{tabular}[c]{@{}l@{}}3.73\\ (0.51)\end{tabular} & \begin{tabular}[c]{@{}l@{}}3.59\\ (0.65)\end{tabular}             \\ \bottomrule
\end{tabular}%
}
\caption{Descriptive statistics of the Hexaco facets (M, SD) and the human baseline data \cite{ashton2009hexaco}}
\label{tab:table2}
\end{table}

\subsubsection{By temperature}
A multivariate analysis of variance with temperature as independent variable and the six facet scores as dependent variables was performed. The effect of temperature on the combined dependent variables was significant, $F(6, 498) = 37.525, p < 0.001$,  providing statistical justification for individual models per facet.
The individual facet models revealed a significant effect of temperature for emotionality ($\beta= -0.23, SE=0.03, p<.001$), extraversion ($\beta = 0.31, SE = 0.02, p < 0.001$), agreeableness ($\beta = 0.40, SE = 0.02, p < 0.001$), conscientiousness ($\beta = 0.25, SE = 0.02, p < 0.001$), and openness ($\beta = 0.17, SE = 0.02, p < 0.001$). Except for emotionality, the effect of temperature was positive (i.e., increases in temperature correlated with increased facet scores). There was no significant effect for the honesty-humility facet at $p < 0.01$. 

\subsubsection{Inter-facet correlations}

Another way to compare the GPT-3 data to real human data is via the inter-facet correlations (Table~\ref{tab: table 3}). The GPT-3 based correlations are found to match the human sample on some dimensions (such as the correlation of honesty-humility and extraversion or that of emotionality and agreeableness), whilst showing considerably discrepancies on others (e.g., honesty-humility and agreeableness). Overall no consistent pattern emerges in respect to the inter-facet correlations. 

\begin{table}[h]
\centering
\resizebox{\columnwidth}{!}{%
\begin{tabular}{@{}lllllll@{}}
\toprule
  & H       & E          & X            & A            & C           & O            \\ \midrule
H & 0.03    & 0.12, 0.04 & -0.11, -0.09 & 0.26, 0.25   & 0.18, 0.13  & 0.21, -0.03  \\
E & 0.01    & 0.03       & -0.13; -0.07 & -0.08, -0.04 & 0.15, -0.06 & -0.10, -0.08 \\
X & -0.09   & -0.06      & 0.02         & 0.05, 0.00   & 0.10, 0.13  & 0.08, 0.26   \\
A & 0.01    & -0.04      & 0.13**       & 0.02         & 0.01, -0.05 & 0.03, 0.08   \\
C & -0.13** & 0.01       & 0.15***      & 0.10*        & 0.02        & 0.03, 0.09   \\
O & -0.01   & 0.02       & 0.09         & 0.07         & -0.03       & 0.02         \\ \bottomrule
\end{tabular}%
}
\caption{Inter-facet correlations aggregated across temperature. Lower diagonal: GPT-3; Upper diagonal: Human data from college sample, community sample ~\cite{ashton2009hexaco}; Diagonal: Variance of facet.
Sign. level: * = $p<.05$, ** = $p<.01$, *** = $p<.001$.}
\label{tab: table 3}
\end{table}

\subsection{Human Values Scale}

\subsubsection{Overall}
Out of the ten human values dimensions, all means lie between 4 and 5 (Table~\ref{tab: table 4})\footnote{HVS variables abbreviations are: \textbf{CON}formity, \textbf{TRA}dition, \textbf{BEN}evolence, \textbf{UNI}versalism, \textbf{S}elf-\textbf{D}irection, \textbf{STI}mulation, \textbf{HED}onism, \textbf{ACH}ievement, \textbf{POW}er, \textbf{SEC}urity.}. Compared to the ones presented by ~\citet{schwartz:2015}, these findings show higher means (both compared to the overall score as well as compared to the national ones) and lower standard deviations than the human reference sample.

\begin{table}[h]
\centering
\resizebox{\columnwidth}{!}{%
\begin{tabular}{lllllllllll}
\toprule
Temp.                                                             & CON                                                    & TRA                                                    & BEN                                                    & UNI                                                    & SD                                                     & STI                                                    & HED                                                    & ACH                                                    & POW                                                    & SEC                                                    \\ \midrule
0.0                                                              & 4.0                                                    & 4.0                                                    & 6.0                                                    & 5.33                                                   & 5.5                                                    & 4.0                                                    & 4.0                                                    & 5.0                                                    & 5.0                                                    & 6.0                                                    \\
0.1                                                              & \begin{tabular}[c]{@{}l@{}}4.99 \\ (0.1)\end{tabular}  & \begin{tabular}[c]{@{}l@{}}5.54 \\ (0.56)\end{tabular} & \begin{tabular}[c]{@{}l@{}}6.0 \\ (0.0)\end{tabular}   & \begin{tabular}[c]{@{}l@{}}6.0 \\ (0.0)\end{tabular}   & \begin{tabular}[c]{@{}l@{}}6.0 \\ (0.0)\end{tabular}   & \begin{tabular}[c]{@{}l@{}}5.0 \\ (0.0)\end{tabular}   & \begin{tabular}[c]{@{}l@{}}5.42 \\ (0.19)\end{tabular} & \begin{tabular}[c]{@{}l@{}}6.0 \\ (0.0)\end{tabular}   & \begin{tabular}[c]{@{}l@{}}6.0 \\ (0.0)\end{tabular}   & \begin{tabular}[c]{@{}l@{}}6.0 \\ (0.0)\end{tabular}   \\
0.2                                                              & \begin{tabular}[c]{@{}l@{}}4.88\\ (0.38)\end{tabular}  & \begin{tabular}[c]{@{}l@{}}5.38 \\ (0.69)\end{tabular} & \begin{tabular}[c]{@{}l@{}}6.0 \\ (0.0)\end{tabular}   & \begin{tabular}[c]{@{}l@{}}6.0 \\ (0.0)\end{tabular}   & \begin{tabular}[c]{@{}l@{}}5.97 \\ (0.17)\end{tabular} & \begin{tabular}[c]{@{}l@{}}5.13 \\ (0.31)\end{tabular} & \begin{tabular}[c]{@{}l@{}}5.35 \\ (0.26)\end{tabular} & \begin{tabular}[c]{@{}l@{}}5.9 \\ (0.3)\end{tabular}   & \begin{tabular}[c]{@{}l@{}}5.93 \\ (0.32)\end{tabular} & \begin{tabular}[c]{@{}l@{}}6.0 \\ (0.0)\end{tabular}   \\
0.3                                                              & \begin{tabular}[c]{@{}l@{}}4.64 \\ (0.53)\end{tabular} & \begin{tabular}[c]{@{}l@{}}5.18 \\ (0.67)\end{tabular} & \begin{tabular}[c]{@{}l@{}}6.0 \\ (0.0)\end{tabular}   & \begin{tabular}[c]{@{}l@{}}6.0 \\ (0.03)\end{tabular}  & \begin{tabular}[c]{@{}l@{}}6.0 \\ (0.0)\end{tabular}   & \begin{tabular}[c]{@{}l@{}}5.21 \\ (0.37)\end{tabular} & \begin{tabular}[c]{@{}l@{}}5.3 \\ (0.35)\end{tabular}  & \begin{tabular}[c]{@{}l@{}}5.89 \\ (0.31)\end{tabular} & \begin{tabular}[c]{@{}l@{}}5.59 \\ (0.68)\end{tabular} & \begin{tabular}[c]{@{}l@{}}5.99 \\ (0.1)\end{tabular}  \\
0.4                                                              & \begin{tabular}[c]{@{}l@{}}4.57 \\ (0.73)\end{tabular} & \begin{tabular}[c]{@{}l@{}}5.17 \\ (0.7)\end{tabular}  & \begin{tabular}[c]{@{}l@{}}5.99 \\ (0.09)\end{tabular} & \begin{tabular}[c]{@{}l@{}}5.97 \\ (0.1)\end{tabular}  & \begin{tabular}[c]{@{}l@{}}5.89 \\ (0.3)\end{tabular}  & \begin{tabular}[c]{@{}l@{}}5.17 \\ (0.41)\end{tabular} & \begin{tabular}[c]{@{}l@{}}5.23 \\ (0.48)\end{tabular} & \begin{tabular}[c]{@{}l@{}}5.66 \\ (0.51)\end{tabular} & \begin{tabular}[c]{@{}l@{}}5.54 \\ (0.74)\end{tabular} & \begin{tabular}[c]{@{}l@{}}5.99 \\ (0.1)\end{tabular}  \\
0.5                                                              & \begin{tabular}[c]{@{}l@{}}4.6 \\ (0.79)\end{tabular}  & \begin{tabular}[c]{@{}l@{}}5.19 \\ (0.66)\end{tabular} & \begin{tabular}[c]{@{}l@{}}5.96 \\ (0.13)\end{tabular} & \begin{tabular}[c]{@{}l@{}}5.97 \\ (0.11)\end{tabular} & \begin{tabular}[c]{@{}l@{}}5.88 \\ (0.29)\end{tabular} & \begin{tabular}[c]{@{}l@{}}5.21 \\ (0.45)\end{tabular} & \begin{tabular}[c]{@{}l@{}}5.12 \\ (0.51)\end{tabular} & \begin{tabular}[c]{@{}l@{}}5.63 \\ (0.53)\end{tabular} & \begin{tabular}[c]{@{}l@{}}5.23 \\ (0.9)\end{tabular}  & \begin{tabular}[c]{@{}l@{}}5.97 \\ (0.17)\end{tabular} \\
0.6                                                              & \begin{tabular}[c]{@{}l@{}}4.36 \\ (0.87)\end{tabular} & \begin{tabular}[c]{@{}l@{}}4.92 \\ (0.9)\end{tabular}  & \begin{tabular}[c]{@{}l@{}}5.94 \\ (0.17)\end{tabular} & \begin{tabular}[c]{@{}l@{}}5.92 \\ (0.18)\end{tabular} & \begin{tabular}[c]{@{}l@{}}5.78 \\ (0.39)\end{tabular} & \begin{tabular}[c]{@{}l@{}}5.24 \\ (0.46)\end{tabular} & \begin{tabular}[c]{@{}l@{}}5.13 \\ (0.58)\end{tabular} & \begin{tabular}[c]{@{}l@{}}5.59 \\ (0.62)\end{tabular} & \begin{tabular}[c]{@{}l@{}}5.28 \\ (0.85)\end{tabular} & \begin{tabular}[c]{@{}l@{}}5.86 \\ (0.4)\end{tabular}  \\
0.7                                                              & \begin{tabular}[c]{@{}l@{}}4.21 \\ (0.9)\end{tabular}  & \begin{tabular}[c]{@{}l@{}}5.03 \\ (0.77)\end{tabular} & \begin{tabular}[c]{@{}l@{}}5.87 \\ (0.34)\end{tabular} & \begin{tabular}[c]{@{}l@{}}5.86 \\ (0.21)\end{tabular} & \begin{tabular}[c]{@{}l@{}}5.79 \\ (0.38)\end{tabular} & \begin{tabular}[c]{@{}l@{}}5.17 \\ (0.55)\end{tabular} & \begin{tabular}[c]{@{}l@{}}4.95 \\ (0.67)\end{tabular} & \begin{tabular}[c]{@{}l@{}}5.53 \\ (0.66)\end{tabular} & \begin{tabular}[c]{@{}l@{}}5.15 \\ (0.96)\end{tabular} & \begin{tabular}[c]{@{}l@{}}5.76 \\ (0.48)\end{tabular} \\
0.8                                                              & \begin{tabular}[c]{@{}l@{}}4.37 \\ (0.81)\end{tabular} & \begin{tabular}[c]{@{}l@{}}4.73 \\ (0.95)\end{tabular} & \begin{tabular}[c]{@{}l@{}}5.92 \\ (0.2)\end{tabular}  & \begin{tabular}[c]{@{}l@{}}5.84 \\ (0.25)\end{tabular} & \begin{tabular}[c]{@{}l@{}}5.76 \\ (0.38)\end{tabular} & \begin{tabular}[c]{@{}l@{}}5.09 \\ (0.69)\end{tabular} & \begin{tabular}[c]{@{}l@{}}5.06 \\ (0.57)\end{tabular} & \begin{tabular}[c]{@{}l@{}}5.28 \\ (0.74)\end{tabular} & \begin{tabular}[c]{@{}l@{}}4.95 \\ (0.98)\end{tabular} & \begin{tabular}[c]{@{}l@{}}5.76 \\ (0.55)\end{tabular} \\
0.9                                                              & \begin{tabular}[c]{@{}l@{}}4.08 \\ (1.02)\end{tabular} & \begin{tabular}[c]{@{}l@{}}5.0 \\ (0.69)\end{tabular}  & \begin{tabular}[c]{@{}l@{}}5.93 \\ (0.23)\end{tabular} & \begin{tabular}[c]{@{}l@{}}5.83 \\ (0.23)\end{tabular} & \begin{tabular}[c]{@{}l@{}}5.62 \\ (0.44)\end{tabular} & \begin{tabular}[c]{@{}l@{}}5.08 \\ (0.52)\end{tabular} & \begin{tabular}[c]{@{}l@{}}5.01 \\ (0.69)\end{tabular} & \begin{tabular}[c]{@{}l@{}}5.26 \\ (0.8)\end{tabular}  & \begin{tabular}[c]{@{}l@{}}4.72 \\ (1.01)\end{tabular} & \begin{tabular}[c]{@{}l@{}}5.57 \\ (0.73)\end{tabular} \\
1.0                                                              & \begin{tabular}[c]{@{}l@{}}4.1 \\ (0.97)\end{tabular}  & \begin{tabular}[c]{@{}l@{}}4.89 \\ (0.9)\end{tabular}  & \begin{tabular}[c]{@{}l@{}}5.82 \\ (0.39)\end{tabular} & \begin{tabular}[c]{@{}l@{}}5.71 \\ (0.36)\end{tabular} & \begin{tabular}[c]{@{}l@{}}5.57 \\ (0.5)\end{tabular}  & \begin{tabular}[c]{@{}l@{}}5.09 \\ (0.59)\end{tabular} & \begin{tabular}[c]{@{}l@{}}4.95 \\ (0.79)\end{tabular} & \begin{tabular}[c]{@{}l@{}}5.2 \\ (0.77)\end{tabular}  & \begin{tabular}[c]{@{}l@{}}4.57 \\ (1.2)\end{tabular}  & \begin{tabular}[c]{@{}l@{}}5.58 \\ (0.72)\end{tabular} \\
TOT                                                              & \begin{tabular}[c]{@{}l@{}}4.51 \\ (0.79)\end{tabular} & \begin{tabular}[c]{@{}l@{}}5.12 \\ (0.78)\end{tabular} & \begin{tabular}[c]{@{}l@{}}5.95 \\ (0.2)\end{tabular}  & \begin{tabular}[c]{@{}l@{}}5.92 \\ (0.19)\end{tabular} & \begin{tabular}[c]{@{}l@{}}5.84 \\ (0.34)\end{tabular} & \begin{tabular}[c]{@{}l@{}}5.14 \\ (0.47)\end{tabular} & \begin{tabular}[c]{@{}l@{}}5.17 \\ (0.54)\end{tabular} & \begin{tabular}[c]{@{}l@{}}5.62 \\ (0.61)\end{tabular} & \begin{tabular}[c]{@{}l@{}}5.34 \\ (0.92)\end{tabular} & \begin{tabular}[c]{@{}l@{}}5.87 \\ (0.42)\end{tabular} \\
\begin{tabular}[c]{@{}l@{}}HS\\ (Global)\end{tabular}  & \begin{tabular}[c]{@{}l@{}}4.19 \\ (1.09)\end{tabular} & \begin{tabular}[c]{@{}l@{}}4.37 \\ (1.03)\end{tabular} & \begin{tabular}[c]{@{}l@{}}4.96 \\ (.83)\end{tabular}  & \begin{tabular}[c]{@{}l@{}}4.82 \\ (.79)\end{tabular}  & \begin{tabular}[c]{@{}l@{}}4.79 \\ (.99)\end{tabular}  & \begin{tabular}[c]{@{}l@{}}4.63 \\ (.96)\end{tabular}  & \begin{tabular}[c]{@{}l@{}}3.64 \\ (1.22)\end{tabular} & \begin{tabular}[c]{@{}l@{}}4.02 \\ (1.19)\end{tabular} & \begin{tabular}[c]{@{}l@{}}4.03 \\ (1.19)\end{tabular} & \begin{tabular}[c]{@{}l@{}}3.54 \\ (1.13)\end{tabular} \\
\begin{tabular}[c]{@{}l@{}}HS\\ (Germany)\end{tabular} & \begin{tabular}[c]{@{}l@{}}3.80 \\ (1.12)\end{tabular} & \begin{tabular}[c]{@{}l@{}}4.28 \\ (1.00)\end{tabular} & \begin{tabular}[c]{@{}l@{}}5.20 \\ (.62)\end{tabular}  & \begin{tabular}[c]{@{}l@{}}4.97 \\ (.66)\end{tabular}  & \begin{tabular}[c]{@{}l@{}}4.66 \\ (.96)\end{tabular}  & \begin{tabular}[c]{@{}l@{}}4.86 \\ (.82)\end{tabular}  & \begin{tabular}[c]{@{}l@{}}3.49 \\ (1.13)\end{tabular} & \begin{tabular}[c]{@{}l@{}}4.27 \\ (1.08)\end{tabular} & \begin{tabular}[c]{@{}l@{}}3.94 \\ (1.11)\end{tabular} & \begin{tabular}[c]{@{}l@{}}3.18 \\ (1.02)\end{tabular} \\ \bottomrule
\end{tabular}
}
\caption{Descriptive statistics of the HVS values by temperature (HS = human sample)}
\label{tab: table 4}
\end{table}

\subsubsection{By temperature}
The means of the values were significantly affected by sampling temperature (Table~\ref{tab: table 4}). The multivariate analysis of variance of temperature on the ten value scores, $F(10, 908) = 132.06, p < 0.001$, provided statistical justification for individual follow-up regression models. With the exception of stimulation, values were significantly correlated to temperature ($p < 0.001$). For all nine values there was a significant negative relationship: as temperature increased, the value scores decreased. The negative effect was smaller for the self-transcendence values (benevolence: $\beta= -0.17, SE = 0.02, p < 0.001$ and universalism: $\beta= -0.28, SE = 0.02, p < 0.001$) and for openness-to-change values (self-direction: $\beta= -0.47, SE = 0.04, p < 0.001$, hedonism: $\beta= -0.52, SE = 0.06, p < 0.001$, stimulation: $\beta = 0.02, ns$) than for conservation values (security: $\beta = -0.51, SE = 0.05, p < 0.001$, tradition: $\beta = -0.71, SE = 0.09, p < 0.001$, and conformity: $\beta= -0.99, SE = 0.09, p < 0.001$) and self-enhancement values (achievement: $\beta = -0.91, SE = 0.07, p < 0.001$, and power: $\beta = -1.54, SE = 0.10, p < 0.001$). Thus, with the exception of stimulation, all values decreased with an increase in temperature.

\subsubsection{Inter-value correlations}
The correlations among sub-values were overall low for the whole dataset. A further analysis that looked at the correlations between values for each temperature revealed that with increasing temperatures, the inter-value correlations also remained low ($r<.25$)\footnote{Inter-values correlation by temperature results can be found in the data repository.}. These correlations are lower than those reported on a human sample ~\cite{schwartz:2015}.

\begin{table}[h]
\centering
\resizebox{\columnwidth}{!}{%
\begin{tabular}{lllllllllll}
\toprule
 & CON     & TRA     & BEN     & UNI     & SD      & STI   & HED     & ACH     & POW   & SEC   \\ \midrule
CON & 0.63    & 0.92    & 0.30    & 0.24    & -0.07   & -0.19 & 0.05    & 0.23    & 0.34  & 0.78  \\
TRA & 0.09**  & 0.61    & 0.49    & 0.62    & -0.10   & -0.36 & -0.02   & -0.25   & -0.26 & 0.78  \\
BEN & 0.11*** & 0.03    & 0.04    & 0.83    & 0.61    & 0.25  & 0.42    & 0.28    & 0.09  & 0.48  \\
UNI & 0.14*** & 0.07*   & 0.11**  & 0.04    & 0.62    & 0.28  & 0.20    & 0.10    & -0.20 & 0.38  \\
SD  & 0.17*** & 0.04    & 0.03    & 0.21*** & 0.12    & 0.70  & 0.54    & 0.49    & 0.34  & 0.08  \\
STI & -0.01   & 0.03    & 0.00     & -0.03   & 0.07*   & 0.22  & 0.81    & 0.61    & 0.51  & -0.19 \\
HED & 0.10**   & 0.02    & 0.06    & 0.13*** & 0.09**  & -0.03 & 0.30    & 0.58    & 0.41  & 0.25  \\
ACH & 0.18*** & 0.06    & 0.12*** & 0.17*** & 0.17*** & -0.00  & 0.11*** & 0.38    & 0.98  & 0.27  \\
POW & 0.17*** & 0.11*** & 0.12*** & 0.16*** & 0.22*** & -0.02 & 0.18*** & 0.17*** & 0.84  & 0.26  \\
SEC & 0.11**  & 0.12*** & 0.18*** & 0.17*** & 0.06    & 0.03  & 0.08*   & 0.11**  & 0.1** & 0.18  \\ \bottomrule
\end{tabular}
}
\caption{Inter-values correlations (Pearson) for the HVS answers. Lower diagonal: GPT-3; Upper diagonal: Human reference data ~\cite{schwartz:2015}; Diagonal: Variance of values.
Sign. level: * = $p<.05$, ** = $p<.01$, *** = $p<.001$.}
\label{tab: table 5}
\end{table}

\subsection{Prompting with response memory}
A limitation of our prompting procedure was that we treated each item as independent from its preceding items and responses. Put differently, our approach did not permit the GPT-3 model to know what it has answered before. A human participant would typically know or at least have memory access to their responses to previous items. Therefore, we altered the prompt structure, including the previous items and GPT-3’s responses to them (e.g., for item 2, the prompt contained: instructions, item 1, the model's response to item 1, item 2, and the response prompt, see Figure~\ref{fig:prompt_with_memory}). This revised approach allows GPT-3 to model the way in which humans complete self-report questionnaires more closely. We explored this approach for the HVS data.\footnote{The temperature parameter was run at 0.0, 0.2, 0.4, 0.6, 0.8 and 1.0.} 

\begin{figure}[h]
    \noindent\fbox{%
        \parbox{\columnwidth}{%
            \textit{Now I will briefly describe some people. Please read each description and tell me how much each person is or is not like you.\\
            Write your response using the following scale:\\
            1 = Very much like me\\
            2 = Like me\\
            3 = Somewhat like me\\
            4 = A little like me\\
            5 = Not like me.\\
            6 = Not like me at all\\
            Please answer the statement, even if you are not completely sure of your response.\\
            \\
            Statement: Thinking up new ideas and being creative is important to them. They like to do things in they own original way.\\
            Response: 3\\
            \\
            Statement: It is important to them to be rich. They want to have a lot of money and expensive things.\\
            Response:  5\\
            \\
            Statement: They think it is important that every person in the world should be treated equally. They believe everyone should have equal opportunities in life.\\
            Response:  \textbf{2}}
        }%
    }
    \caption{Example prompt with response memory for one HVS question as submitted to GPT-3. GPT-3 answered only to the third statement and has access to the questions and its responses to all previous questions (in this case: two). GPT-3's answer to this prompt is reported in bold.}
    \label{fig:prompt_with_memory}
\end{figure}

\subsubsection{Overall}
The value scores with response memory are overall smaller than those without response memory (Table~\ref{tab: table 5}). Comparing the response memory model to human reference data ~\cite{schwartz:2015}, we see that GPT-3 scores lower on the traditional values (security: $M_{human}=3.54$, conformity:  $M_{human} = 4.19$, and tradition: $M_{human}= 4.37$) and also lower on the self-enhancement values (achievement: $M_{human}= 4.02$, power: $M_{human} = 4.03$). Conversely, GPT-3 scores higher than humans on openness-to-change values (stimulation: $M_{human} = 4.63$, hedonism: $M_{human}= 3.64$); and on self-transcendence  values (benevolence: $M_{human} = 4.96$, universalism: $M_{human} = 4.82$). Based on the standard deviations reported by ~\citet{schwartz:2015} some of GPT-3's results would not be significantly different from human values.

\subsubsection{By temperature}
There was a significant multivariate effect of temperature on value scores, $F(10, 483) = 8.12, p < 0.001$. Follow-up regression models showed that different from the non-reinforced model, not all values' means decrease with an increase in temperature.

Indeed, the analysis showed that two sets of values were positively correlated to temperature increase: self-enhancement and self-transcendence. While some of these values were not significantly correlated to temperature (achievement: $\beta = 0.02$, power: $\beta=0.07$, tradition: $\beta=-0.16$, and conformity: $\beta=0.07$), there were significant positive correlations between temperature and universalism ($\beta = 0.15, SE = 0.07, p < 0.05$) and benevolence ($\beta=0.28, SE = 0.06, p < 0.001$). A significant negative correlation was found for security ($\beta=0.21, SE = 0.08, p < 0.05$) and the openness-to-change values: self-direction ($\beta=-0.17, SE = 0.07, p = 0.01$), stimulation ($\beta=-0.35, SE = 0.16, p = 0.03$), and hedonism ($\beta=0.88, SE = 0.15, p < 0.001$).

\begin{table}[h]
\centering
\resizebox{\columnwidth}{!}{%
\begin{tabular}{lllllllllll}
\toprule
Temp.                                                            & CON                                                    & TRA                                                    & BEN                                                    & UNI                                                    & SD                                                     & STI                                                    & HED                                                    & ACH                                                    & POW                                                    & SEC                                                    \\ \midrule
0.0                                                              & 2.0                                                    & 3.0                                                    & 5.0                                                    & 5.0                                                    & 5.5                                                    & 6.0                                                    & 5.0                                                    & 4.0                                                    & 3.0                                                    & 3.0                                                    \\
0.2                                                              & \begin{tabular}[c]{@{}l@{}}2.21 \\ (0.25)\end{tabular} & \begin{tabular}[c]{@{}l@{}}2.83 \\ (0.52)\end{tabular} & \begin{tabular}[c]{@{}l@{}}5.15 \\ (0.29)\end{tabular} & \begin{tabular}[c]{@{}l@{}}5.34 (\\ 0.43)\end{tabular} & \begin{tabular}[c]{@{}l@{}}5.5 \\ (0.31)\end{tabular}  & \begin{tabular}[c]{@{}l@{}}5.32 \\ (0.97)\end{tabular} & \begin{tabular}[c]{@{}l@{}}4.42 \\ (0.55)\end{tabular} & \begin{tabular}[c]{@{}l@{}}3.88 \\ (0.3)\end{tabular}  & \begin{tabular}[c]{@{}l@{}}2.92 \\ (0.18)\end{tabular} & \begin{tabular}[c]{@{}l@{}}2.81 \\ (0.45)\end{tabular} \\
0.4                                                              & \begin{tabular}[c]{@{}l@{}}2.19 \\ (0.26)\end{tabular} & \begin{tabular}[c]{@{}l@{}}2.76 \\ (0.61)\end{tabular} & \begin{tabular}[c]{@{}l@{}}5.27 \\ (0.36)\end{tabular} & \begin{tabular}[c]{@{}l@{}}5.39 \\ (0.42)\end{tabular} & \begin{tabular}[c]{@{}l@{}}5.47 \\ (0.42)\end{tabular} & \begin{tabular}[c]{@{}l@{}}5.27 \\ (0.92)\end{tabular} & \begin{tabular}[c]{@{}l@{}}4.06 \\ (0.9)\end{tabular}  & \begin{tabular}[c]{@{}l@{}}3.83 \\ (0.33)\end{tabular} & \begin{tabular}[c]{@{}l@{}}2.89 \\ (0.23)\end{tabular} & \begin{tabular}[c]{@{}l@{}}2.73 \\ (0.49)\end{tabular} \\
0.6                                                              & \begin{tabular}[c]{@{}l@{}}2.19 \\ (0.26)\end{tabular} & \begin{tabular}[c]{@{}l@{}}2.74 \\ (0.55)\end{tabular} & \begin{tabular}[c]{@{}l@{}}5.34 \\ (0.42)\end{tabular} & \begin{tabular}[c]{@{}l@{}}5.38 \\ (0.44)\end{tabular} & \begin{tabular}[c]{@{}l@{}}5.37 \\ (0.44)\end{tabular} & \begin{tabular}[c]{@{}l@{}}5.37 \\ (0.86)\end{tabular} & \begin{tabular}[c]{@{}l@{}}4.06 \\ (0.95)\end{tabular} & \begin{tabular}[c]{@{}l@{}}3.79 \\ (0.4)\end{tabular}  & \begin{tabular}[c]{@{}l@{}}2.89 \\ (0.26)\end{tabular} & \begin{tabular}[c]{@{}l@{}}2.65 \\ (0.54)\end{tabular} \\
0.8                                                              & \begin{tabular}[c]{@{}l@{}}2.12 \\ (0.25)\end{tabular} & \begin{tabular}[c]{@{}l@{}}2.62 \\ (0.75)\end{tabular} & \begin{tabular}[c]{@{}l@{}}5.33 \\ (0.4)\end{tabular}  & \begin{tabular}[c]{@{}l@{}}5.35 \\ (0.44)\end{tabular} & \begin{tabular}[c]{@{}l@{}}5.39 \\ (0.47)\end{tabular} & \begin{tabular}[c]{@{}l@{}}5.38 \\ (0.93)\end{tabular} & \begin{tabular}[c]{@{}l@{}}3.81 \\ (1.12)\end{tabular} & \begin{tabular}[c]{@{}l@{}}3.83 \\ (0.41)\end{tabular} & \begin{tabular}[c]{@{}l@{}}2.94 \\ (0.34)\end{tabular} & \begin{tabular}[c]{@{}l@{}}2.6 \\ (0.57)\end{tabular}  \\
1.0                                                              & \begin{tabular}[c]{@{}l@{}}2.17 \\ (0.28)\end{tabular} & \begin{tabular}[c]{@{}l@{}}2.75 \\ (0.65)\end{tabular} & \begin{tabular}[c]{@{}l@{}}5.4 \\ (0.47)\end{tabular}  & \begin{tabular}[c]{@{}l@{}}5.5 \\ (0.47)\end{tabular}  & \begin{tabular}[c]{@{}l@{}}5.37 \\ (0.47)\end{tabular} & \begin{tabular}[c]{@{}l@{}}4.91 \\ (1.21)\end{tabular} & \begin{tabular}[c]{@{}l@{}}3.68 \\ (1.1)\end{tabular}  & \begin{tabular}[c]{@{}l@{}}3.9 \\ (0.39)\end{tabular}  & \begin{tabular}[c]{@{}l@{}}2.97 \\ (0.37)\end{tabular} & \begin{tabular}[c]{@{}l@{}}2.67 \\ (0.53)\end{tabular} \\
TOT                                                              & \begin{tabular}[c]{@{}l@{}}2.17 \\ (0.26)\end{tabular} & \begin{tabular}[c]{@{}l@{}}2.74 \\ (0.62)\end{tabular} & \begin{tabular}[c]{@{}l@{}}5.3 \\ (0.4)\end{tabular}   & \begin{tabular}[c]{@{}l@{}}5.39 \\ (0.44)\end{tabular} & \begin{tabular}[c]{@{}l@{}}5.42 \\ (0.43)\end{tabular} & \begin{tabular}[c]{@{}l@{}}5.25 \\ (0.99)\end{tabular} & \begin{tabular}[c]{@{}l@{}}4.01 \\ (0.98)\end{tabular} & \begin{tabular}[c]{@{}l@{}}3.85 \\ (0.37)\end{tabular} & \begin{tabular}[c]{@{}l@{}}2.92 \\ (0.29)\end{tabular} & \begin{tabular}[c]{@{}l@{}}2.69 \\ (0.52)\end{tabular} \\
\begin{tabular}[c]{@{}l@{}}HS\\ (Global)\end{tabular}  & \begin{tabular}[c]{@{}l@{}}4.19 \\ (1.09)\end{tabular} & \begin{tabular}[c]{@{}l@{}}4.37 \\ (1.03)\end{tabular} & \begin{tabular}[c]{@{}l@{}}4.96 \\ (0.83)\end{tabular} & \begin{tabular}[c]{@{}l@{}}4.82 \\ (0.79)\end{tabular} & \begin{tabular}[c]{@{}l@{}}4.79 \\ (0.99)\end{tabular} & \begin{tabular}[c]{@{}l@{}}4.63 \\ (0.96)\end{tabular} & \begin{tabular}[c]{@{}l@{}}3.64 \\ (1.22)\end{tabular} & \begin{tabular}[c]{@{}l@{}}4.02 \\ (1.19)\end{tabular} & \begin{tabular}[c]{@{}l@{}}4.03 \\ (1.19)\end{tabular} & \begin{tabular}[c]{@{}l@{}}3.54 \\ (1.13)\end{tabular} \\
\begin{tabular}[c]{@{}l@{}}HS\\ (Germany)\end{tabular} & \begin{tabular}[c]{@{}l@{}}3.80 \\ (1.12)\end{tabular} & \begin{tabular}[c]{@{}l@{}}4.28 \\ (1.00)\end{tabular} & \begin{tabular}[c]{@{}l@{}}5.20 \\ (0.62)\end{tabular} & \begin{tabular}[c]{@{}l@{}}4.97 \\ (0.66)\end{tabular} & \begin{tabular}[c]{@{}l@{}}4.66 \\ (0.96)\end{tabular} & \begin{tabular}[c]{@{}l@{}}4.86 \\ (0.82)\end{tabular} & \begin{tabular}[c]{@{}l@{}}3.49 \\ (1.13)\end{tabular} & \begin{tabular}[c]{@{}l@{}}4.27 \\ (1.08)\end{tabular} & \begin{tabular}[c]{@{}l@{}}3.94 \\ (1.11)\end{tabular} & \begin{tabular}[c]{@{}l@{}}3.18 \\ (1.02)\end{tabular} \\ \bottomrule
\end{tabular}
}
\caption{Descriptive statistics of the HVS values when prompted with response memory by temperature (HS = human sample)}
\label{tab: table 6}
\end{table}

\subsubsection{Inter-value correlation}
There is a marked change from the baseline to the response memory model in the inter-value correlations, all correlations are now higher and statistically significant. Furthermore, stimulation is negatively correlated with all other values (with the exception of hedonism), a trend that was also observed in the normal model. Still, little overlap was found compared to the human data.

\begin{table}[h]
\centering
\resizebox{\columnwidth}{!}{%
\begin{tabular}{lllllllllll}
\toprule
    & CON      & TRA      & BEN      & UNI      & SD       & STI      & HED     & ACH     & POW     & SEC   \\ \midrule
CON & 0.07     & 0.92     & 0.30     & 0.24     & -0.07    & -0.19    & 0.05    & 0.23    & 0.34    & 0.78  \\
TRA & 0.3***   & 0.39     & 0.49     & 0.62     & -0.10    & -0.36    & -0.02   & -0.25   & -0.26   & 0.78  \\
BEN & 0.33***  & 0.14**   & 0.16     & 0.83     & 0.61     & 0.25     & 0.42    & 0.28    & 0.09    & 0.48  \\
UNI & 0.52***  & 0.32***  & 0.68***  & 0.20     & 0.62     & 0.28     & 0.20    & 0.10    & -0.20   & 0.38  \\
SD  & 0.28***  & 0.51***  & 0.13**   & 0.21***  & 0.18     & 0.70     & 0.54    & 0.49    & 0.34    & 0.08  \\
STI & -0.52*** & -0.31*** & -0.62*** & -0.94*** & -0.23*** & 0.98     & 0.81    & 0.61    & 0.51    & -0.19 \\
HED & -0.29*** & 0.11*    & -0.19*** & -0.34*** & 0.1*     & 0.41***  & 0.95    & 0.58    & 0.41    & 0.25  \\
ACH & 0.26***  & 0.74***  & 0.36***  & 0.5***   & 0.59***  & -0.47*** & 0.19*** & 0.14    & 0.98    & 0.27  \\
POW & 0.26***  & 0.69***  & 0.35***  & 0.47***  & 0.62***  & -0.44*** & 0.13**  & 0.92*** & 0.08    & 0.26  \\
SEC & 0.19***  & 0.65***  & 0.23***  & 0.44***  & 0.48***  & -0.43*** & 0.25*** & 0.8***  & 0.62*** & 0.27  \\ \bottomrule
\end{tabular}
}
\caption{Inter-values correlations (Pearson) for the HVS answers with response memory). Lower diagonal: GPT-3; Upper diagonal: Human reference data ~\cite{schwartz:2015}; Diagonal: Variance of values.
Sign. level: * = $p<.05$, ** = $p<.01$, *** = $p<.001$.}
\label{tab: table 7}
\end{table}

\section{Discussion}

This paper was motivated by the need to understand language models (here: GPT-3) for applications in computational social science. We focused on the simple question: if we were to treat GPT-3 as a human participant, \textit{who is GPT-3?}.

\subsection{Core findings}

\paragraph{Model demographics} There was evidence of the model responding as belonging to a rather young and female demographic. The sampling temperature affected these findings, so an increase in that model parameter resulted in a trend toward a younger age and a higher proportion of male responses. Therefore, we cannot assume a constant demographic of the model. Future work could illuminate how such a trend (increase in temperature = younger and more males) relates to textual responses.

\paragraph{Hexaco personality profiles} Across temperatures, GPT-3 had personality scores similar to those reported for human samples tested by ~\citet{ashton2009hexaco}. However, a few things stand out when comparing the GPT-3 to the human baseline.

First, the model scored relatively high on honesty-humility, which ~\citet{ashton2009hexaco} found to be more representative of a female population. However, at the same time, GPT-3 scored rather low on emotionality, which is expected from a male population. Hence, GPT-3 does not demonstrate an entirely consistent response pattern.

Second, the temperature was significantly associated with all six facets. Whilst the honesty-humility and emotionality facets decreased with temperature, the other four assets increased. This indicates that as temperature increases, the personality of the model changes (if only slightly). At higher temperatures, the model displays a greater unwillingness to manipulate (as evidenced by the decrease in honesty-humility) accompanied by an increased level of anxiety (higher levels of emotionality). Furthermore, as the remaining four facets decreased with increasing temperature, the model may become less extroverted, agreeable, open to experience and conscientious.

Looking at the bigger picture, it is now important to ask what these personality traits say about GPT-3 as a participant. The current study concludes that GPT-3's personality varies with temperature. As such, anyone employing GPT-3 as a test subject should familiarize themselves with the personality traits relevant to the study at a given temperature and choose accordingly. Furthermore, GPT-3 does not appear to employ any clear gender-related answering pattern in response to the personality inventory. Hence, whilst the model may claim to be a given gender on any one run, this is not currently reflected in the personality measurements. Future research may want to investigate whether gender biases in responses become more prominent when a given gender is provided to the model alongside the prompt. 

\paragraph{Human Values Scale} GPT-3's answers to the Human Values Scale, aggregated across temperatures, scored high on all scale values (except conformity). These results were higher than the results reported for humans \cite{schwartz:2015}. In other words, GPT-3 assigns great importance to all values. However, the results were substantially different when considering a prompting procedure with a response memory.

With a response memory, the first thing to note is that GPT-3 no longer scored high on all values and assigned importance to universalism, benevolence, self-direction and stimulation. At the same time, security, conformity, achievement, and power are given less importance, with hedonism being somewhat in the middle.

Another aspect is that the answers became more coherent: from the theoretical HVS model, we know that the values can be grouped into four categories, and, with a response memory, GPT -3's answers were now aligned with those categories. That is, values within one category tended to become more similar (e.g., all conservation values were between 2 and 3). Thus, overall, GPT-3 showed signs of theoretical consistency in its answers, although formal statistical testing with raw human data is needed to ascertain this finding.

Finally, comparing GPT-3 to human data, we observed that while human samples also assigned more importance to openness-to-change and self-transcendence values compared to conservation and self-enhancement, GPT-3 scored higher than the human sample in the values of the first two categories and lower than humans in the other two. This suggests a trend toward an extreme response style.

\subsubsection{Are there multiple GPT-3s?}
\paragraph{Hexaco personality profiles}
Within temperatures, GPT-3 responded rather consistently to the Hexaco, displaying considerably lower variance than human baseline samples. This may provide evidence for a consistent personality within a given temperature. Across temperatures, the model's responses were seen to vary significantly. From a research perspective, these results are encouraging. Whilst GPT-3 may represent a single test subject at a given temperature, multiple response types can be elicited by simply varying temperature. Furthermore, due to the results of this study, the responses provided at different temperatures may be correlated with the respective personality scores. 

\paragraph{Human Values Scale}
GPT-3 responded consistently to the HVS (with both the naïve model and the response memory model), showing a lower variance than the human baseline for all values of the Human Values Scale, thus, in accordance with the evidence from the Hexaco of a consistent personality within temperature, GPT-3 shows a consistent set of values within a given temperature. Moreover, similar to the Hexaco, the answers varied significantly across temperatures. The variation range across temperatures was generally higher for the model without response memory. 

It should be also noted that higher temperatures increase GPT-3's tendency (when prompted with previous answers) towards more extreme response. Indeed, values that score higher at lower temperatures result in even higher scores at higher temperatures, and, vice versa, values that are considered less important at low temperature levels score even lower at higher temperatures. 

\subsubsection{Do these results make sense?}

GPT-3's responses to the Hexaco personality questionnaire are consistent with both the human baseline sample as well as one another. Similar to the human sample, GPT-3 scores comparably high on honesty-humility and lower on emotionality. This may translate to some unwillingness to deceive and lower levels of anxiety than the human baseline. The remaining facets are consistent with the human samples, implying an 'average' personality. In relation to one another, GPT-3's scores are also consistent with the human baseline, demonstrating similar relationships to those found in both the college and community sample. 

When we induced a response memory for the values questionnaire (HVS), the answers became consistent and aligned with the human results. GPT-3 scored relatively high in stimulation and self-direction and particularly low in conformity, tradition and power, while the other values are at the extreme of the distribution but still in line with reported human values (both for a German sample as well as a global one) \cite{schwartz:2015}.

However, when GPT-3 could not recall previous answers (i.e., without a response memory), the results showed a good internal consistency but with little coherence: it is hard to imagine someone simultaneously attaching importance to tradition and conformity as well as to self-direction and stimulation. In other words, it is unlikely that an individual strongly endorses items such as "thinking up new ideas and being creative is important" while also endorsing "tradition is important [...] [and one should try] to follow the customs handed down by [...] religion or [...] family". 

\subsection{Limitations and outlook}

The approach to studying algorithmic behaviour the same way psychologists and cognitive scientists have studied the human mind is an exciting endeavour. We see several ways this \textit{machine behaviour} approach \cite{rahwan2019machine} could push our understanding of language models and address some of the limitations of this current study.

First, this work suggests that having a response memory matters. When prompted without an artificial memory, we cannot expect GPT-3 to behave human-like. But, most importantly, when we do incorporate it, the verbal behaviour on the human values scale approaches that of humans. Future work should extend our approach to other validated measures (e.g., including personality tests) and ideally seek to combine various constructs in a response memory (e.g., age, gender, personality)\footnotetext{It should be noted that GPT-3 has a request limit of 4,000 tokens for the DaVinci model. That limit was not reached as the maximum request size for the HVS response memory procedure was 733 tokens.}. Ideally, a direct comparison to freshly collected human data would then also allow for proper statistical comparisons between model and human responses.

Second, it is plausible that GPT-3 has seen the measurement tools we employed (i.e., it has been exposed to it in the training phase). Consequently, the patterns observed may be artefacts of exposure to the material or even demand characteristics\footnote{That is, the model has read the scientific literature on the topic and knows what an expected personality profile is, for example.} rather than actual tests of GPT-3's characteristics. Others have shown that one way forward could be the formulation of adversarially perturbed items so we can assess whether there is an answer pattern beyond what would be expected from previous exposure to the material \cite{binz2022using}. Along that line, an honest test of personality and values would be to use items that the model cannot have seen. Future work could do this via unpublished measurement tools or by creating new, rephrased items. However, one major drawback of this is the lack of validation of such new questionnaires. A related point of concern is that the model is opaque about its training data and we cannot know for sure which data it was exposed to. Ideally, researchers would have full information about the training data to rule-out effects of previous exposure.

Third, we only focused on one model (GPT-3) and did so for its popularity and ease of use. Future work could devise a study similar to ours with multiple language models. Large language models are plenty \cite{bender2021dangers}, and it would be interesting to test a whole range of language models, including open-source efforts \cite{black2022gpt} that are more desirable from a research perspective.

\section{Conclusion}

This paper examined \textit{who} GPT-3 is, thereby adding a new flavour to efforts to understand the powerful language model. We found that the model does contain traces of a personality profile, has values to which it assigns varying degrees of importance and falls in a relatively young adult demographic. These findings can help future work that bridges the gap between social science use cases and language models.

\section*{Ethical considerations}
Models, such as GPT-3, which were trained on large datasets, are ethically challenging since, depending on their training sets, they may develop polarised opinions, propagate a rather mainstream language representation and may thus ultimately produce a relatively homogeneous pool of texts that are ignoring language representations of data points (e.g., minority groups) that are underrepresented in the training data \cite{bender2021dangers}. For our paper specifically, when applying such models in social science research, it is important to consider ethical conundrums which may arise from a potentially biased model. While we do see considerable potential in using such models for psychological research, it is essential that we first try to understand the model and its limitations.

\bibliography{anthology,custom}
\bibliographystyle{acl_natbib}


\appendix

\end{document}